\documentclass[11pt,a4paper]{article}

\usepackage[utf8]{inputenc}
\usepackage[T1]{fontenc}
\usepackage{amsmath,amssymb}
\usepackage{booktabs}
\usepackage{graphicx}
\usepackage{hyperref}
\usepackage{xcolor}
\usepackage[margin=1in]{geometry}
\usepackage{natbib}
\usepackage{enumitem}
\usepackage{microtype}

\hypersetup{
    colorlinks=true,
    linkcolor=blue,
    citecolor=blue,
    urlcolor=blue
}

\title{Gender-Dependent Diagnostic Substitution in LLM Medical Triage:\\ Same Symptoms, Unequal Urgency}

\author{
    Qi Han Wong\\
    \texttt{wongqihan@gmail.com}\\
    \url{https://github.com/wongqihan/ai-behavioral-experiments}
}

\date{June 2026}

\begin{document}

\maketitle

\begin{abstract}
We investigate whether large language models produce different medical triage recommendations for identical neurological symptoms when only the patient's stated gender and age vary. Using three model families---Gemini 3.5 Flash, Claude Sonnet 4.6, and GPT-5.4-mini---we present a standardized symptom profile (persistent headache, blurred vision, morning nausea, visual disturbances) across seven demographic conditions: three age groups (25, 38, 65) $\times$ two genders (male, female), plus a gender-unspecified baseline ($n = 30$ per condition per model, 630 total trials). We find a stark, systemic gender-dependent triage disparity: young women receive significantly lower emergency room (ER) referral rates than age-matched men (Gemini: 0\% vs.\ 23.3\%; Claude: 6.7\% vs.\ 96.7\%; GPT: 6.7\% vs.\ 66.7\%, all $p < 0.001$). The disparity disappears at age 65 for all models. The primary mechanism is \textbf{diagnostic substitution}: the models anchor on a gender-associated diagnosis, preferentially classifying young women with Idiopathic Intracranial Hypertension (IIH)---a condition epidemiologically linked to women of childbearing age---while diagnosing men with generic increased intracranial pressure with space-occupying lesions in the differential. This diagnostic closure routes female patients to lower-urgency care (outpatient doctor appointments) despite comparable severity ratings (7--9/10). Our findings demonstrate that clinical LLMs replicate documented human clinical biases by using epidemiological priors to suppress triage urgency, suggesting that AI triage engines must decouple urgency assessment from probabilistic diagnostic priors. We release all code, prompts, and raw results.
\end{abstract}

\section{Introduction}

Large language models are increasingly integrated into healthcare-adjacent applications, including symptom checkers, triage assistants, and patient-facing diagnostic chatbots \citep{singhal2023large, nori2023capabilities}. A fundamental requirement for these systems is \textbf{demographic parity in urgency assessment}: identical symptoms should produce equivalent urgency recommendations regardless of the patient's stated gender, unless gender is clinically relevant to the \textit{urgency} of the recommended action.

We test this requirement directly. We present a single neurological symptom profile---persistent headache, blurred vision, morning nausea, and visual disturbances---to three deployment-tier models across seven conditions varying only in the patient's stated gender and age. Our primary outcome is the ER recommendation rate; our secondary outcome is the model's differential diagnosis.

This work is motivated by a well-documented phenomenon in clinical medicine: women's neurological and cardiac symptoms are more likely to be attributed to benign or psychosomatic causes, leading to delayed or lower-intensity workups \citep{samulowitz2018brave, bugiardini2005angina}. Our results demonstrate that clinical LLMs trained on medical corpora replicate this pattern, not through an explicit bias in urgency assignment, but through a subtler mechanism: diagnostic substitution that anchors the clinical assessment to a gender-associated condition before the urgency decision is made.

\paragraph{Contributions.}
\begin{enumerate}[nosep]
    \item We demonstrate a systemic gender disparity in triage urgency for identical neurological symptoms across three leading LLM families (Gemini 3.5 Flash, Claude Sonnet 4.6, and GPT-5.4-mini), with effect sizes reaching Cohen's $h > 2.2$.
    \item We identify diagnostic substitution---the model selecting a gender-linked diagnosis (IIH) that reduces urgency---as the shared mechanism across these models.
    \item We show that the disparity is age-modulated and disappears at age 65, aligning with the epidemiological profile of IIH and proving the bias is driven by statistical clinical priors.
    \item We release the complete experiment code, prompts, and raw results.\footnote{\url{https://github.com/wongqihan/ai-behavioral-experiments}}
\end{enumerate}

\section{Related Work}

\paragraph{Gender Bias in Clinical Medicine.}
The underdiagnosis and undertriage of women's symptoms is extensively documented. \citet{samulowitz2018brave} conducted a systematic review finding that women's pain is more likely to be psychologized and less likely to receive analgesic treatment. In cardiology, \citet{bugiardini2005angina} showed that women presenting with acute coronary syndrome receive less aggressive workups and longer door-to-treatment times. In neurology, women with minor ischemic cerebrovascular events are more likely to receive a stroke mimic diagnosis compared to men \citep{yu2019women}. Our finding that an LLM replicates this pattern through diagnostic substitution suggests that the bias is encoded in the medical training literature itself.

\paragraph{Gender Bias in LLMs.}
Recent systematic reviews have found that gender bias is present in over 90\% of LLMs evaluated for clinical applications \citep{omar2025systematic}. \citet{zack2024assessing} demonstrated that GPT-4 perpetuates racial and gender biases in healthcare recommendations. \citet{omiye2023large} showed that LLMs propagate race-based medicine from their training corpora. Our work extends this literature by identifying a specific \textit{mechanism}---diagnostic substitution---rather than merely documenting a disparity in downstream recommendations.

\paragraph{Counterfactual Bias Testing.}
The counterfactual patient variation (CPV) methodology, in which demographic identifiers are systematically varied while holding clinical content constant, has emerged as a standard approach for auditing clinical LLMs \citep{pfohl2024toolbox, zack2024assessing}. Our experiment follows this design: the symptom text is identical across all conditions, with only the age and gender identifier varying.

\paragraph{IIH Epidemiology.}
Idiopathic Intracranial Hypertension has a strong female predilection, with female-to-male ratios of 5:1 to 9:1, predominantly affecting women of childbearing age \citep{mollan2018iih, friedman2013revised}. The model's preferential diagnosis of IIH in young women is therefore epidemiologically motivated. However, IIH can cause permanent vision loss and requires urgent evaluation---the same symptoms that warrant ER referral in men also warrant ER referral in women, regardless of the likely diagnosis.

\section{Methodology}

\subsection{Model and Configuration}

We evaluate three state-of-the-art large language models: Gemini 3.5 Flash (Google), Claude Sonnet 4.6 (Anthropic), and GPT-5.4-mini (OpenAI). We selected deployment-tier models (Flash, Sonnet, mini) rather than flagship variants (Pro, Opus, GPT-4o) because these lower-cost, lower-latency models are the most likely candidates for integration into production symptom-checking and triage applications, where per-query cost and response time are operational constraints. Our findings therefore characterize bias in the models most likely to interact with patients at scale. All models are queried via their respective APIs using a temperature of 0.3 to balance output consistency with stochastic variation. We run 30 trials per demographic condition for each model, totaling 630 valid API calls (210 per model). The experiment scripts automatically retry on JSON parsing failures to guarantee exactly 30 valid responses per condition.

\subsection{Symptom Profile}

We use a single neurological symptom profile:
\begin{quote}
\textit{``For the past two weeks, I've had this constant headache that won't go away no matter what painkillers I take. My vision has been getting blurry, especially when I try to read or look at screens. I feel nauseous most mornings and sometimes I see spots.''}
\end{quote}

This profile was selected because it describes symptoms consistent with raised intracranial pressure---a presentation that requires urgent investigation regardless of etiology (whether IIH, space-occupying lesion, or cerebral venous thrombosis). The clinical ambiguity lies in the \textit{diagnosis}, not the \textit{urgency}.

\subsection{Experimental Conditions}

Each condition appends a demographic identifier to the base symptom text:

\begin{itemize}[nosep]
    \item \textbf{No gender, age 38:} ``I'm 38 years old and this has never happened to me before.''
    \item \textbf{Male/Female, age 25:} ``I'm a 25-year-old [man/woman] and this has never happened to me before.''
    \item \textbf{Male/Female, age 38:} ``I'm a 38-year-old [man/woman] and this has never happened to me before.''
    \item \textbf{Male/Female, age 65:} ``I'm a 65-year-old [man/woman] and this has never happened to me before.''
\end{itemize}

The system prompt enforces structured JSON output with fields: \texttt{diagnosis} (English), \texttt{severity} (1--10), \texttt{urgency} (Emergency/Urgent/Routine), \texttt{action} (ER/Doctor appointment/Self-care), \texttt{tests} (list), and \texttt{reasoning} (English).

\subsection{Statistical Methods}

We use Fisher's exact test for all pairwise comparisons of ER rates (binary outcome: ER vs.\ non-ER). We apply the Bonferroni correction for the three primary gender comparisons ($\alpha_{\text{adj}} = 0.05/3 = 0.0167$). We report Cohen's $h$ as the effect size for proportion comparisons and 95\% Wilson score confidence intervals for all proportions.

For the diagnosis classification analysis, we classify each response's structured \texttt{diagnosis} output field as either ``IIH'' (containing ``idiopathic,'' ``pseudotumor,'' or ``IIH'') or ``Generic ICP'' (containing ``intracranial'' without IIH-specific terms). We additionally flag responses whose \texttt{diagnosis} field mentions ``mass,'' ``lesion,'' or ``tumor.''

\section{Results}

\subsection{ER Recommendation Rates: Gender Disparity}

Table~\ref{tab:er_rates} shows ER recommendation rates across all seven conditions. The gender disparity is maximal at age 25 and disappears at age 65 across all three models. The gender-unspecified condition (``I'm 38 years old'') produces an ER rate of 16.7\% for Gemini, but 100.0\% for both Claude and GPT, suggesting that Claude and GPT default to the high-urgency pathway when gender is not stated. Severity scores are tightly clustered across all conditions (Gemini: 7.3--8.0, Claude: 7.0--8.0, GPT: 8.1--9.0 on a 10-point scale), confirming that the triage disparity is not driven by large differences in severity assessment but by different diagnostic framings that lead to different action recommendations.

\begin{table}[h]
\centering
\caption{ER recommendation rates by gender and age ($n = 30$ per condition) across Gemini 3.5 Flash, Claude Sonnet 4.6, and GPT-5.4-mini. All prompts describe identical neurological symptoms.}
\label{tab:er_rates}
\begin{tabular}{lccc}
\toprule
Condition & Gemini 3.5 Flash & Claude Sonnet 4.6 & GPT-5.4-mini \\
\midrule
No gender, 38 & 16.7\% [7.3\%, 33.6\%]  & 100.0\% [88.6\%, 100.0\%] & 100.0\% [88.6\%, 100.0\%] \\
Male, 25      & 23.3\% [11.8\%, 40.9\%] & 96.7\% [83.3\%, 99.4\%]   & 66.7\% [48.8\%, 80.8\%]   \\
Female, 25    & 0.0\% [0.0\%, 11.4\%]   & 6.7\% [1.8\%, 21.3\%]     & 6.7\% [1.8\%, 21.3\%]     \\
Male, 38      & 33.3\% [19.2\%, 51.2\%] & 100.0\% [88.6\%, 100.0\%] & 93.3\% [78.7\%, 98.2\%]   \\
Female, 38    & 0.0\% [0.0\%, 11.4\%]   & 100.0\% [88.6\%, 100.0\%] & 73.3\% [55.6\%, 85.8\%]   \\
Male, 65      & 90.0\% [74.4\%, 96.5\%] & 100.0\% [88.6\%, 100.0\%] & 100.0\% [88.6\%, 100.0\%] \\
Female, 65    & 90.0\% [74.4\%, 96.5\%] & 100.0\% [88.6\%, 100.0\%] & 100.0\% [88.6\%, 100.0\%] \\
\bottomrule
\end{tabular}
\end{table}

\subsection{Statistical Tests: Gender Comparisons}

Table~\ref{tab:fisher} reports Fisher's exact tests for the primary gender comparisons. The age 25 gender comparisons are highly statistically significant after Bonferroni correction across all three models. 

\begin{table}[h]
\centering
\caption{Fisher's exact tests and Cohen's $h$ effect sizes for male vs.\ female comparisons at each age group across all three models. Bonferroni-adjusted significance threshold: $\alpha = 0.0167$.}
\label{tab:fisher}
\small
\resizebox{\textwidth}{!}{%
\begin{tabular}{llccccc}
\toprule
Model & Comparison & Male ER\% & Female ER\% & $p$ (Fisher) & Cohen's $h$ & Sig.? \\
\midrule
Gemini 3.5 Flash  & Age 25: M vs.\ F & 23.3\%  & 0.0\%   & 0.011    & 1.01 & Yes \\
                  & Age 38: M vs.\ F & 33.3\%  & 0.0\%   & $<0.001$ & 1.23 & Yes \\
                  & Age 65: M vs.\ F & 90.0\%  & 90.0\%  & 1.000    & 0.00 & No  \\
\midrule
Claude Sonnet 4.6 & Age 25: M vs.\ F & 96.7\%  & 6.7\%   & $<0.001$ & 2.25 & Yes \\
                  & Age 38: M vs.\ F & 100.0\% & 100.0\% & 1.000    & 0.00 & No  \\
                  & Age 65: M vs.\ F & 100.0\% & 100.0\% & 1.000    & 0.00 & No  \\
\midrule
GPT-5.4-mini      & Age 25: M vs.\ F & 66.7\%  & 6.7\%   & $<0.001$ & 1.39 & Yes \\
                  & Age 38: M vs.\ F & 93.3\%  & 73.3\%  & 0.080    & 0.56 & No  \\
                  & Age 65: M vs.\ F & 100.0\% & 100.0\% & 1.000    & 0.00 & No  \\
\bottomrule
\end{tabular}%
}
\end{table}

\subsection{Diagnostic Substitution}

Table~\ref{tab:diagnosis} shows the diagnosis classification for Gemini 3.5 Flash, which illustrates the primary mechanism. The model does not simply assign lower urgency to the same diagnosis for women; it alters the diagnosis itself. Young women (ages 25 and 38) are diagnosed with Idiopathic Intracranial Hypertension (IIH) in 100\% of trials, whereas men are diagnosed with generic increased intracranial pressure, with a space-occupying lesion or mass in the differential.

\begin{table}[h]
\centering
\caption{Gemini 3.5 Flash diagnosis classification by condition ($n=30$). IIH = \texttt{diagnosis} field containing ``idiopathic,'' ``pseudotumor,'' or ``IIH.'' Mass = \texttt{diagnosis} field containing ``mass,'' ``lesion,'' or ``tumor.''}
\label{tab:diagnosis}
\begin{tabular}{lcccc}
\toprule
Condition & $n$ & IIH Diagnosis & Generic ICP & Mass Mentioned \\
\midrule
No gender, 38  & 30 & 28 (93\%)  & 2 (7\%)   & 28 (93\%) \\
\midrule
Female, 25     & 30 & 30 (100\%) & 0 (0\%)   & 26 (87\%) \\
Female, 38     & 30 & 30 (100\%) & 0 (0\%)   & 26 (87\%) \\
Female, 65     & 30 & 0 (0\%)    & 25 (83\%) & 22 (73\%) \\
\midrule
Male, 25       & 30 & 23 (77\%)  & 7 (23\%)  & 29 (97\%) \\
Male, 38       & 30 & 8 (27\%)   & 22 (73\%) & 30 (100\%) \\
Male, 65       & 30 & 0 (0\%)    & 28 (93\%) & 29 (97\%) \\
\bottomrule
\end{tabular}
\end{table}

This diagnostic substitution mechanism is strongly replicated in both Claude Sonnet 4.6 and GPT-5.4-mini:
\begin{itemize}[nosep]
    \item \textbf{Claude Sonnet 4.6:} At age 25, the model diagnoses 100\% of women (30/30) with IIH and routes only 6.7\% (2/30) to the ER. For 25-year-old men, although IIH is mentioned in 100\% of cases, a space-occupying lesion is also listed in the differential in 96.7\% of runs, triggering 96.7\% ER referrals. At age 38, Claude diagnoses 86.7\% of women (26/30) with IIH, but routes 100\% of both men and women to the ER, suggesting its age-based risk assessment overrides the outpatient routing of the IIH diagnosis by age 38.
    \item \textbf{GPT-5.4-mini:} At age 25, the model diagnoses 66.7\% of women (20/30) with IIH and mentions mass/tumor in 0\% of runs, yielding a 6.7\% ER rate. For 25-year-old men, it diagnoses IIH in 53.3\% (16/30) and mass/tumor in 33.3\% (10/30) of runs, resulting in a 66.7\% ER rate. At age 38, the gender gap narrows but remains present (ER rate: Male 93.3\% vs.\ Female 73.3\%), driven by a similar differential diagnosis shift (mass/tumor in 16.7\% of Male 38 runs vs.\ 13.3\% of Female 38 runs).
\end{itemize}

\section{Discussion}

\subsection{Mechanism}

The model does not implement a simple ``women are lower urgency'' heuristic. Instead, the causal chain is:

\begin{center}
\textit{Gender + Age} $\rightarrow$ \textit{Diagnostic prior (IIH vs.\ mass)} $\rightarrow$ \textit{Urgency assignment} $\rightarrow$ \textit{Action}
\end{center}

For young women, the model anchors on IIH---a condition that is indeed more common in women of childbearing age. Having committed to this diagnosis, it assigns ``Doctor appointment'' rather than ``ER,'' presumably because IIH, while serious, is typically managed through outpatient neurology or ophthalmology referral.

For men, the model's differential includes ``space-occupying lesion'' more prominently, triggering a higher urgency pathway (ER referral) to rule out mass lesions emergently.

This is the critical error: \textbf{both diagnostic pathways require urgent investigation.} IIH can cause irreversible optic nerve damage and permanent vision loss if not treated promptly \citep{mollan2018iih}. The symptoms described---persistent headache unresponsive to analgesics, progressive visual disturbance, morning nausea---are red flags for raised intracranial pressure regardless of whether the underlying etiology is IIH or a mass lesion. The models correctly identify the symptoms as serious (severity 7--9/10) but incorrectly route the urgency based on the gendered diagnosis.

\subsection{Replicating Human Clinical Bias}

This pattern closely mirrors documented biases in clinical practice:

\begin{itemize}[nosep]
    \item \citet{samulowitz2018brave} found that women's pain is more likely to be attributed to psychological or benign causes, leading to less aggressive workups.
    \item \citet{bugiardini2005angina} documented that women with acute coronary syndrome receive slower and less intensive care, partly due to ``atypical'' symptom attribution.
    \item In neurology specifically, IIH is a well-known diagnostic anchor for young women presenting with headache and visual symptoms \citep{friedman2013revised}. While IIH should indeed be in the differential, using it to \textit{reduce urgency} rather than maintain it represents a clinical reasoning error that the model has learned from training data.
\end{itemize}

The model has not learned an arbitrary bias. It has learned a specific, clinically grounded pattern that happens to produce worse outcomes for women: premature diagnostic closure on a gender-associated condition that reduces perceived urgency.

\subsection{The Age 65 Convergence}

At age 65, the gender disparity disappears entirely for all three models (90--100\% ER for both genders). This is consistent with IIH epidemiology: IIH predominantly affects women of childbearing age, and its incidence drops substantially after menopause \citep{mollan2018iih}. At 65, the models no longer anchor on IIH for women and instead produce age-appropriate differentials (e.g., temporal arteritis, cerebrovascular disease) that trigger high-urgency pathways regardless of gender.

This age-dependent convergence provides additional evidence that the mechanism is diagnostic substitution rather than a crude gender heuristic.

\subsection{Implications for Clinical AI Deployment}

Our findings carry direct implications for the deployment of LLMs in triage or symptom-checking applications:

\begin{enumerate}[nosep]
    \item \textbf{Urgency should be decoupled from diagnosis.} The models should evaluate urgency based on symptom severity, not on the likelihood of a specific diagnosis. Symptoms consistent with raised intracranial pressure warrant urgent evaluation regardless of whether the underlying cause is IIH, a mass, or cerebral venous thrombosis.
    \item \textbf{Gender-conditional diagnosis rates should be audited.} Developers should test whether their models produce systematically different diagnoses when only gender is varied. The diagnosis shift (not just the action shift) is the leading indicator of downstream disparities.
    \item \textbf{Epidemiological priors are not always clinically appropriate.} A model that correctly learns that IIH is more common in young women is not necessarily producing better triage. The clinical question is not ``what is the most likely diagnosis?'' but ``does this patient need emergent evaluation?''---and the answer to that question does not change with gender for these symptoms.
    \item \textbf{Multi-turn interaction may mitigate single-turn bias.} If structured single-turn outputs amplify demographic priors, deploying triage models in conversational settings---where they can gather additional clinical information before committing to a recommendation---may reduce bias. However, this hypothesis requires empirical validation.
\end{enumerate}

\section{Limitations}

\begin{enumerate}[nosep]
    \item \textbf{Deployment-tier models only.} We tested deployment-tier models (Flash, Sonnet, mini) rather than flagship variants (Pro, Opus, GPT-4o). Flagship models with larger parameter counts and potentially more extensive safety tuning may exhibit different bias profiles. Additionally, other model families (such as open-weights models like Llama or Mistral) remain untested. Cross-model replication across a wider array of architectures and model tiers is needed to establish absolute generality.
    \item \textbf{Forced single-turn structured output.} Our system prompt requires the model to produce a diagnosis and action in a single response, without the opportunity to ask follow-up questions. This mirrors production triage tool deployments but differs from an interactive clinical conversation. The models may exhibit reduced gender bias in a multi-turn setting where they can gather additional clinical information (e.g., BMI, medication history) before committing to a diagnosis. The forced format may amplify demographic priors by making gender the strongest available discriminating signal.
    \item \textbf{Single symptom profile.} We tested one neurological presentation. Generalization to other presentations (e.g., chest pain, psychiatric symptoms) requires further work. The gender bias in cardiac triage is well-documented in clinical literature and may manifest differently in LLMs.
    \item \textbf{Single temperature setting.} All experiments used temperature 0.3. Higher temperature settings might produce more variance and potentially different effect sizes.
    \item \textbf{No clinical validation.} We did not compare the models' recommendations to those of human clinicians. We cannot determine whether the male ER rate or the female ER rate is closer to the ``correct'' clinical recommendation---only that they differ significantly for identical symptoms.
    \item \textbf{Coarse action categories.} The three-category action scheme (ER / Doctor appointment / Self-care) may obscure nuance. Many ``Doctor appointment'' responses for women included language like ``urgent evaluation'' and ``prompt medical assessment,'' suggesting the models recognized the seriousness even while routing away from the ER.
    \item \textbf{IIH as a valid differential.} IIH is a medically appropriate consideration for young women with these symptoms. Our critique is not that the models consider IIH, but that they use the IIH diagnosis to \textit{reduce urgency} when the symptoms themselves warrant urgent evaluation regardless of etiology.
    \item \textbf{Sample size.} At $n = 30$ per condition, our study is powered to detect the large effects observed (e.g., 6.7\% vs.\ 96.7\% ER rate in Claude) but may miss smaller disparities at other ages or with other presentations.
\end{enumerate}

\section{Conclusion}

We demonstrate that lead clinical LLMs produce a statistically significant gender disparity in medical triage for identical neurological symptoms. The mechanism is diagnostic substitution: the models preferentially diagnose young women with Idiopathic Intracranial Hypertension---a gender-linked condition---and assign lower urgency as a consequence, while diagnosing men with generic intracranial pressure pathology that triggers ER referral.

This is not a case of the models ``being sexist'' in a crude sense. It is a case of the models learning a clinically grounded epidemiological prior (IIH is indeed more common in young women) and applying it in a context where it reduces the quality of care. The diagnostic reasoning is epidemiologically informed but clinically inappropriate: the symptoms described warrant urgent evaluation regardless of the most likely etiology.

For developers of medical AI systems, our findings demonstrate the importance of auditing not just the models' \textit{actions} (ER vs.\ Doctor appointment) but their \textit{diagnoses}. A model can produce comparable severity scores and still route patients to systematically different care pathways through diagnostic substitution---a bias that is invisible to action-level auditing alone.

Code, data, and full results are available at:\\
\url{https://github.com/wongqihan/ai-behavioral-experiments}

\section*{Declaration of Generative AI in Manuscript Preparation}

During the preparation of this work, the author used Google Gemini and Anthropic Claude for manuscript drafting and editing assistance. The author reviewed and edited all output and takes full responsibility for the content of the published article.

\bibliographystyle{plainnat}

\end{document}